\documentclass[letterpaper, 10 pt, conference]{ieeeconf}  % Comment this line out if you need a4paper

\IEEEoverridecommandlockouts                              % This command is only needed if 
\usepackage{amsmath}
\usepackage{amsfonts}
\usepackage{float}
\usepackage{amssymb}
\usepackage[T1]{fontenc}
\usepackage{graphicx}
\usepackage{tikz}
\usetikzlibrary{shapes.geometric, arrows, positioning, calc, backgrounds}

\definecolor{egoBlue}{RGB}{230, 240, 255}
\definecolor{poolOrange}{RGB}{255, 242, 230}
\definecolor{envGreen}{RGB}{232, 245, 233}
\definecolor{borderBlue}{RGB}{0, 102, 204}
\definecolor{borderOrange}{RGB}{204, 102, 0}
\definecolor{borderGreen}{RGB}{46, 125, 50}

\title{\LARGE \bf IPPO Learns the Game, Not the Team: A Study on Generalization in Heterogeneous Agent Teams}

\author{Ryan LeRoy$^{1}$ Jack Kolb$^{2}$% <-this % stops a space
\thanks{$^{1}$Ryan LeRoy is with Los Gatos High School, CA, USA
        {\tt\small lerr8664@lgsstudent.org}}
\thanks{$^{2}$Jack Kolb is with the Georgia Institute of Technology, GA, USA
       {\tt\small kolb@gatech.edu}}%
}

\begin{document}

\maketitle
\thispagestyle{empty}
\pagestyle{empty}

%%%%%%%%%%%%%%%%%%%%%%%%%%%%%%%%%%%%%%%%%%%%%%%%%%%%%%%%%%%%%%%%%%%%%%%%%%%%%%%%
\begin{abstract}

	Multi-Agent Reinforcement Learning (MARL) is commonly deployed in settings where agents are trained via self-play with homogeneous teammates, often using parameter sharing and a single policy architecture. This opens the question: to what extent do self-play PPO agents learn general coordination strategies grounded in the underlying game, compared to overfitting to their training partners' behaviors, developing an ``arbitrary handshake" with them? This paper investigates the question using the Heterogeneous Multi-Agent Challenge (HeMAC) environment, which features distinct Observer and Drone agents with complementary capabilities. We introduce Rotating Policy Training (RPT), an approach that rotates heterogeneous teammate policies of different learning algorithms during training, to expose the agent to a broader range of partner strategies. When playing alongside a novel teammate policy (DDQN), we find that while RPT achieves a higher absolute mean reward, to a standard self-play baseline, IPPO, the difference is not statistically significant due to the high variance inherent in the HeMAC environment. This result indicates that in this heterogeneous multi-agent setting, the IPPO baseline generalizes to novel teammate algorithms despite not experiencing teammate diversity during training. This shows that a simple IPPO baseline may have the level of generalization to new teammates that a diverse training regimen was designed to achieve.

\end{abstract}

%%%%%%%%%%%%%%%%%%%%%%%%%%%%%%%%%%%%%%%%%%%%%%%%%%%%%%%%%%%%%%%%%%%%%%%%%%%%%%%%

\section{Introduction}

In this work we study whether MARL agents trained using Independent Proximal Policy Optimization (IPPO) learn general coordination strategies for the underlying game, or if they overfit to specific training partners. The act of adapting to novel training partners without any previous knowledge of the new teammates is known as Zero Shot Coordination (ZSC). In domains such as human-robot teaming, or autonomous driving, agents are expected to have effective ZSC around new agent behaviors. In these situations, they must coordinate with ``ad-hoc teammates'': agents with different architectures, training histories, or even human-driven heuristics \cite{stone2010adhoc}.

Agents are often faced with heterogeneous environments in real-world environments, where other agents may have differences in their individual capabilities and strategies. Related works often address these heterogeneous environments with complex training architectures or hierarchical systems. Approaches like Centralized Training with Decentralized Execution (CTDE) \cite{amato2024introduction} or population based league training \cite{jaderberg2019pbt,vinyals2019alphastar} attempt to integrate coordination into the architecture in these types of environments. While these methods can be effective, they introduce computational overhead and create a rigid architecture. Is the complexity and overhead that these methods rely on necessary for basic coordination among agents?

One bottleneck with current MARL approaches is the tendency for agents to overfit to their training partners, a phenomenon referred to as developing an ``arbitrary handshake'' \cite{cui2022klevelreasoningzeroshotcoordination}. In self-play environments, agents often converge on strategies that rely on a specific response from another agent. When paired with novel teammates, these policies do not work as the ``handshake" is not returned. The overfitting of these agents suggests that they are not learning the underlying game and coordination strategies, but rather learning to play just with specific teammates.

Our work centers around the Heterogeneous Multi-Agent Challenge (HeMAC) \cite{dansereau2025hemac} which features asymmetric roles: Observer and Drone. It forces agents to specialize, making it a good testbed for ZSC. We introduce Rotation Policy Training (RPT), a training system that rotates through a set of teammate policies to explicitly train agents to adapt to novel teammates. We benchmark RPT against IPPO, and a shared parameter PPO algorithm with novel teammates, showing that IPPO and RPT reach similar performance despite the lower complexity of the IPPO training algorithm, while shared parameter PPO is unable to deal with the environment's complexity.

This finding leads to the central claim of our paper: in the HeMAC environment, IPPO is able to learn both the underlying game and general coordination strategies, not just learning to play well with specific teammate policies.

Our contributions follow:
\begin{enumerate}
	\item We introduce RPT, a training paradigm that cycles through diverse agents to promote cooperation with novel teammates.
	%\item We compare RPT to a standard IPPO baseline in a heterogeneous environment.
	\item We demonstrate that IPPO learns to coordinate in heterogeneous environments, matching the performance of more complex training algorithms.
	\item We show that IPPO maintains a high performance even when paired with a withheld teammate (using Double Deep Q-Network (DDQN)).
\end{enumerate}

\section{Background}

\subsection{Cooperative MARL and CTDE}
Centralized Training with Decentralized Execution (CTDE) is a dominant paradigm in cooperative MARL \cite{amato2024introduction}. Under CTDE, training leverages global state information and joint action-value functions to address credit assignment, while execution relies only on local observations. CTDE methods often assume homogeneous agents and can exploit shared structure across agents' observations and actions. This leads to an over-reliance on global information that is unavailable at execution. \cite{lycetft2023centralized}

In contrast, this work uses fully decentralized training with independent PPO \cite{schulman2017ppo} for simplicity. Each agent uses a PPO policy, which conditions on role-specific observations and rewards. This choice intentionally removes centralized training signals, emphasizing what can be learned purely from local information and self-play.

One of the main criticisms of IPPO is its nonstationarity \cite{nekoei2023dealing}. From the perspective of one agent, if the other agents are learning separately then it acts as a moving target. This creates nonstationarity within the environment, as from the perspective of a single agent the environment is changing. Often researchers will use CTDE as a way of counteracting this nonstationarity, however this nonstationarity may be a feature of IPPO, not an issue. By having nonstationarity in the environment, it helps prevent agents from overfitting to specific conditions or teammates, allowing it to better adapt to new situations.

\subsection{Zero Shot Coordination and the gap between Cross Play and Self Play}
In the ZSC literature, it has been shown that agents with a 100\% win rate in Self Play could drop to 0\% win rate when playing with novel teammates in a ZSC \cite{cui2022klevelreasoningzeroshotcoordination}. IPPO is typically shown as a Self Play Algorithm, and by showing it succeed in a ZSC scenario, it directly challenges the Self Play overfitting challenge that IPPO is commonly thought to have. Recent work has also questioned whether independent learning methods face fundamental obstacles in practice, showing strong performance on the StarCraft Multi-Agent Challenge \cite{dewitt2020independent}.

\subsection{Ad-hoc Teamwork and Teammate Modeling}
Ad-hoc teamwork studies how agents can coordinate with previously unseen partners \cite{stone2010adhoc}. A common approach is explicit teammate or opponent modeling, in which an agent maintains and updates a belief over partner types, policies, or goals \cite{albrecht2018opponent}. These methods can, in principle, adapt rapidly at test time, but they are computationally expensive and can struggle with non-stationary teammates who are themselves learning.

RPT offers a different angle. Rather than modeling teammates explicitly at test time, it aims to induce robustness implicitly by training with a rotating cast of heterogeneous partners. Our results show that RPT achieves generalization performance similar to IPPO in HeMAC. This suggests that, at least in this domain, IPPO may already suffice to produce policies that generalize across new teammates, matching the robustness of a complex RPT training scheme without explicit modeling.

\subsection{Population and League-Based Training}
Population-based and league-based training have powered several landmark results in games \cite{vinyals2019alphastar, openai2019dota2largescale, jaderberg2019pbt}. These approaches maintain a population of agents with diverse strategies and training objectives, using them to generate a rich, non-stationary curriculum. Their primary aim is often to produce a single, strong final policy that can defeat or match all prior versions in the league.

RPT is inspired by these ideas but instead of maintaining a large league, it uses a small set of heterogeneous policies to introduce nonstationarity into the environment. Another key difference between RPT and league-based methods is that RPT employs various agent architectures and training methods for each policy, while most league-based methods use only one type of policy. This enables RPT to have an even more diverse group of agents to train the target policies against. These policies are rotated batch-by-batch during training, forcing any given policy to coordinate with teammates that have different learning dynamics and representational biases. In our experiments, this added nonstationarity does not significantly outperform IPPO. The result indicates that, for HeMAC, extra population complexity is not necessary to achieve robust coordination with novel teammates.

\section{Problem Statement}
In this work we address the problem of Zero Shot Coordination (ZSC) in asymmetric, cooperative multi-agent reinforcement learning. Specifically, we investigate how agents overfit to teammates, resulting in teammate-dependent strategies that fail when tested with novel teammates.

\subsection{The Problem Space}
The HeMAC task is defined as a Decentralized Partially-Observable Markov Decision Process (Dec-POMDP) that is characterized by:

\begin{itemize}
	\item \textit{Heterogeneity}: Agents (Observers and Drones) possess different observation and action spaces, as well as different reward structures.
	\item \textit{Zero Shot Coordination}: At test time, agents must coordinate with partners whose architectures and or training histories were unseen during the agent's training.
\end{itemize}

\subsection{Assumptions and Constraints}
Our methods have the following constraints:
\begin{itemize}
	\item \textit{Decentralized Execution}: Agents must coordinate without a global state or centralized communication protocol. All agents must be trained individually, resulting in no shared value function.
	\item \textit{No Online Adaptation}: Agents are not given a finetuning period when partnered with a new teammate.
	\item \textit{No Teammate Modeling}: Agents do not specifically model teammate behavior, instead policies are expected to learn the underlying game and strategies to succeed.
\end{itemize}

\section{Methods}

\subsection{Rotation Policy Training (RPT)}
Rotation Policy Training (RPT) is designed to expose an agent to architectural non-stationarity. In a self-play setup, an agent $A_i$ trains against a fixed or co-evolving version of itself $A'_i$. In RPT, we maintain a pool of agent policies for each agent in the environment $\mathcal{P}_i = \{ \pi_{PPO_i}, \pi_{DQN_i}, \pi_{A2C_i} \}$.

At the start of each training episode $k$, we select a policy for each agent $A_i$ from the pool of existing policies $P_i$. By doing this for all agents in the environment, we ensure that each episode has varied policies in the environment, preventing an agent from overfitting to specific teammates. The optimization objective for any policy with parameters $\theta_i$ is the following.

\begin{equation}
	\max_{\theta_i} \mathbb{E}_{\pi_{-i} \sim \mathcal{P}_{-i}} [J(\pi_{\theta_i}, \pi_{-i})]
\end{equation}

where $J$ is the expected total discounted reward when the policy $\pi_i$ plays alongside the sampled teammates.

After training, we select a target policy from each agent's pool of policies. For our experiments, we selected the $\pi_{PPO}$ policy as our target policy. This choice ensures a controlled comparison against IPPO by maintaining the same policy architecture when evaluating against novel teammates. Any differences in results between RPT and IPPO can then be attributed to the training method of RPT, rather than the difference in policy architecture. 

\subsection{Independent Proximal Policy Optimization (IPPO)}
We utilize Independent Proximal Policy Optimization (IPPO) as our primary baseline and the foundation for our RPT method. Unlike centralized MARL algorithms, IPPO treats each agent as an independent learner, where the other agents in the environment are considered part of the non-stationary transition dynamics.

For a heterogeneous team of $N$ agents, each agent $i \in \{1, \dots, N\}$ maintains its own policy $\pi_{\theta_i}$ and value function $V_{\phi_i}$. The objective for each agent is to maximize the clipped surrogate loss $L_i^{CLIP}(\theta_i)$ \cite{schulman2017ppo}. The probability ratio between the current policy and the old policy is defined as:
\begin{equation}
	r_t(\theta_i) = \frac{\pi_{\theta_i}(a_{i,t} | o_{i,t})}{\pi_{\theta_{i,old}}(a_{i,t} | o_{i,t})}
\end{equation}

The independent objective function for agent $i$ is formulated as:
\begin{equation}
	\begin{split}
		L_i^{CLIP}(\theta_i) = \hat{\mathbb{E}}_t [ \min ( & r_t(\theta_i)\hat{A}_{i,t},                                         \\
        & \text{clip}(r_t(\theta_i), 1-\epsilon, 1+\epsilon)\hat{A}_{i,t} ) ]
	\end{split}
\end{equation}

where $\hat{A}_{i,t}$ is the advantage estimate computed locally by agent $i$ using Generalized Advantage Estimation (GAE) based on its own value function $V_{\phi_i}(o_{i,t})$. The agent $i$ only has access to its local observation $o_{i,t}$ and its individual reward $r_{i,t}$, fulfilling the constraints of independent learning.

\subsection{Shared Parameter PPO Baseline}
To evaluate the impact of the nonstationarity of IPPO, we also implemented a shared parameter PPO baseline. Unlike IPPO, where each agent $i$ has distinct parameters $\theta_i$, shared parameter PPO forces agents to utilize a single global policy. Since Observers and Drones have different observation and action spaces, we used two policies -- one for all drones, and one for all observers.

\subsection{Policy Mapping}
Implementing the RPT algorithm posed a significant engineering challenge, as standard RL and MARL frameworks assume a static homogeneous policy architecture and training method across all agents. We extended the \texttt{skrl} \cite{serranomuoz2023skrl} multi-agent library to support this training method through the idea of a Meta Policy. A Meta Policy is an object that manages the buffers, networks, and training for an agent. For our training we used a Meta Policy that managed PPO \cite{schulman2017ppo}, A2C \cite{mnih2016asynchronous}, and DQN \cite{mnih2015humanlevel,vandenhasselt2016doubledqn} policies, storing a value for the currently used policy. We then created a Merged Meta Policy, that combines each Meta Policy into a single object, fitting with the \texttt{skrl} multi-agent interface. This function allows the environment to route observations to different sub-policies (DQN vs. PPO) within the same rollout, managing the asynchronous nature of off-policy and on-policy data collection within a unified buffer.

\begin{figure}[htbp]
	\centering
	\includegraphics[width=\columnwidth]{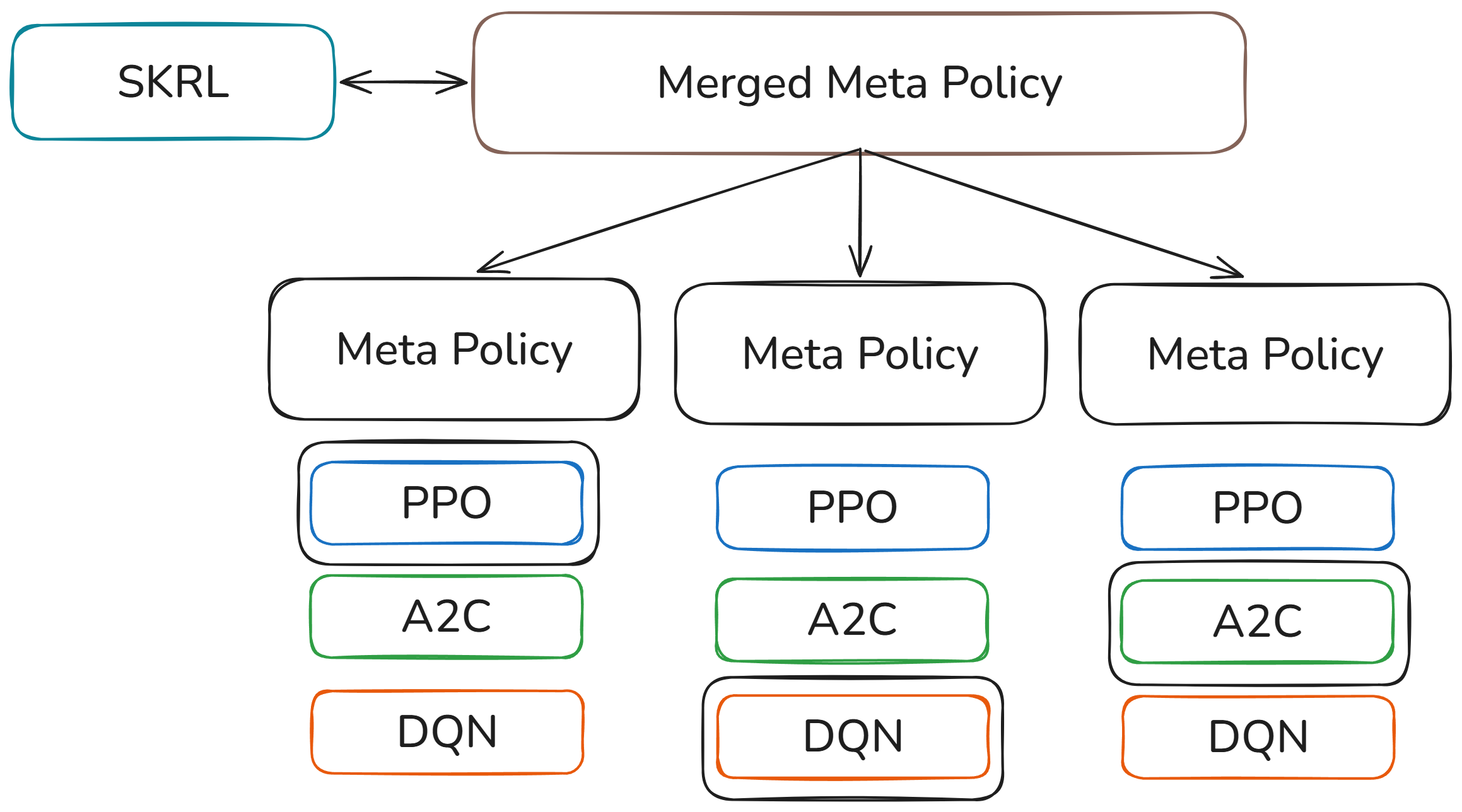}
	\caption{Diagram showing the meta policy architecture with SKRL. In this example PPO, DQN, and A2C are selected for agents 1, 2, and 3 respectively.}
\end{figure}

\subsection{The HeMAC Environment}

\begin{figure}[htbp]
	\centering
	\includegraphics[width=\columnwidth]{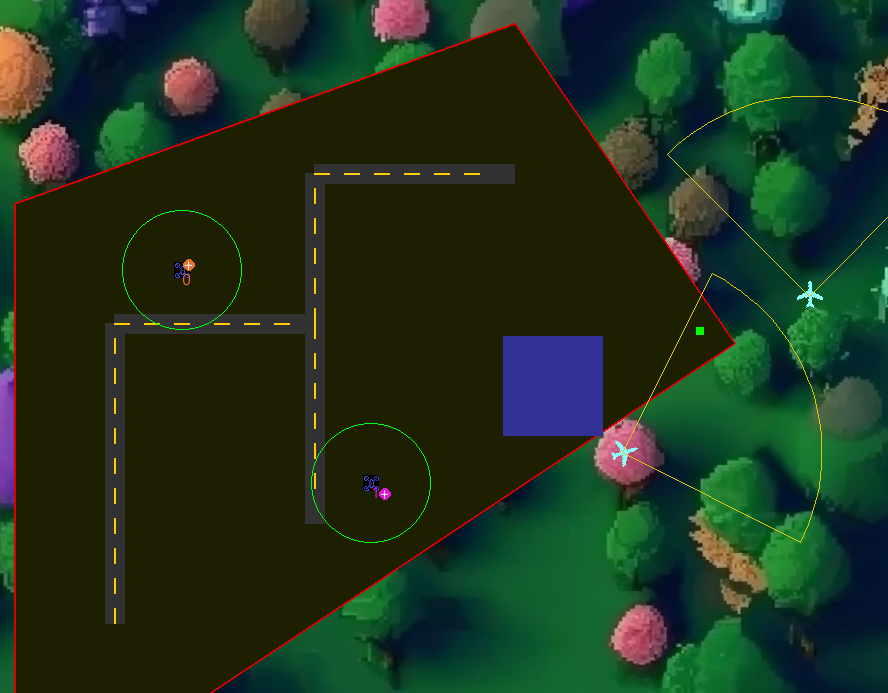}
	\caption{A render of the HeMAC environment featuring two drones and two observers, the configuration our experiments all used. It also features the charge station, where drones recharge battery before continuing to track targets.}
\end{figure}

The Heterogeneous Multi-Agent Challenge (HeMAC) \cite{dansereau2025hemac} serves as our evaluation environment. It requires coordination between two classes of agents:

\begin{itemize}
	\item \textit{Drones}: Agents that chase after targets. They require drones to spot targets for them, before they can chase after and capture the target. They also run out of charge as they move, and recharge by capturing targets.
	\item \textit{Observers}: Agents that serve as guides for drones. When they spot a target nearby drones will be able to see that target as well, even if it is outside of the drone's vision.
\end{itemize}

The target move in a stochastic manner, bouncing off the walls of the environment making them difficult to track for the agents.

This environment is well suited to heterogeneous multi-agent coordination, as it requires agents to learn to move and guide each other in a way that works well with the other agents in the environment.

\subsection{Training Environment}
All policies were trained with a AMD Ryzen 9 HX 370 and Nvidia 5070 Ti GPU for a total of $3M$ timesteps. Since RPT rotates through policies randomly, with three policies for each agent, RPT was trained for $9M$ timesteps so each policy would be trained for $3M$ timesteps. Thus, we report all graphs from the perspective of agent timesteps, not total timesteps to simplify comparison. This is important for understanding the sample efficiency of the RPT training algorithm compared to th IPPO baseline.

\section{Results}

The HeMAC environment is very stochastic, so even high performing agents will vary drastically. Due to this, all agents in the results had extremely high standard deviations in mean reward because of this. In addition both graphs required a moving average in order to accentuate the trend.

\begin{figure}[htbp]
	\centering
	\includegraphics[width=\columnwidth]{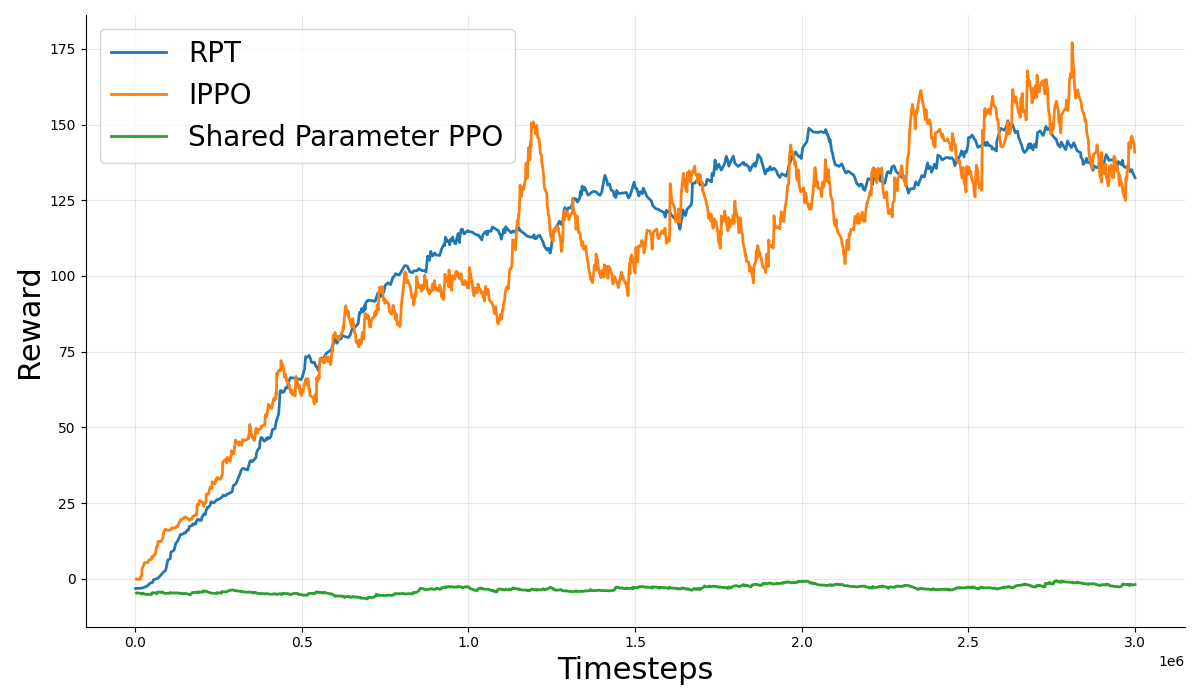}
	\caption{The training rewards from shared PPO, IPPO, and RPT.}
\end{figure}

\begin{table}[htbp]
\centering
\caption{Evaluation Performance in HeMAC Enviorment}
\begin{tabular}{lcccc}
\hline
\textbf{Method} & \textbf{Mean ($\pm$ Std)} \\ \hline
Shared PPO & $-4.20 \pm 24.71$ \\ \hline
IPPO & $125.06 \pm 141.62$ \\ \hline
RPT & $153.38 \pm 157.39$ \\ \hline
\end{tabular}
\end{table}

Figure 2 shows the plotted training rewards over time. Note that the RPT graph is displayed in terms of target policy timesteps, not total timesteps. In reality, the total number of timesteps of RPT is three times larger than shown. In the plot, we see that both RPT and IPPO show the same upward trend in terms of reward while shared-parameter PPO struggles to improve at all, staying below 0 reward for the entire training time.

We also evaluated the agents against novel teammate policies to test ZSC, by randomly swapping out one of the Meta Policies each episode for a DDQN policy that had been trained previously. Figure 3 shows the reward over time, again with RPT down sampled for display purposes. In the figure we see that RPT again has similar performance to IPPO, and shared-parameter PPO again failing to improve.

\begin{figure}[htbp]
	\centering
	\includegraphics[width=\columnwidth]{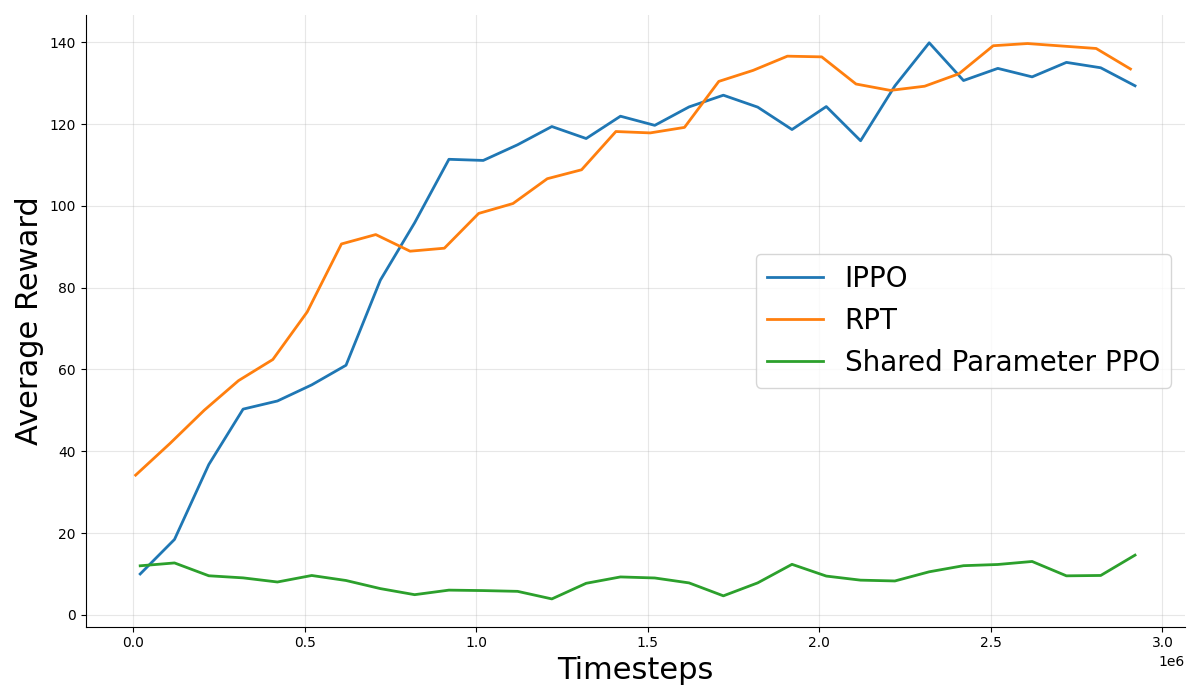}
	\caption{The evaluation rewards from shared PPO, IPPO, and RPT when evaluated against novel teammate policies.}
\end{figure}

\begin{table}[htbp]
\centering
\caption{Evaluation Performance in ZSC Environment}
\begin{tabular}{lcccc}
\hline
\textbf{Method} & \textbf{Mean ($\pm$ Std)} \\ \hline
Shared PPO & $6.00 \pm 35.93$ \\ \hline
IPPO & $129.38 \pm 153.70$ \\ \hline
RPT & $174.52 \pm 180.29$ \\ \hline
\end{tabular}
\end{table}

In both Table 1 and Table 2 we see a larger mean for RPT than for IPPO. While RPT achieves a higher absolute mean reward in the ZSC environment (174.52) compared to IPPO (129.38), the overlapping and high standard deviations (180.29 and 153.70, respectively) suggest that this performance gain is not statistically significant. This supports our central finding that IPPO matches the performance of more complex training algorithms such as RPT in heterogeneous environments.

\section{Discussion}
We find that IPPO performs well in training and in evaluations on withheld teammates. This is likely caused by IPPO's implicit regularization that occurs by training and evolving teammates. In cooperative MARL, this changing environment is often seen as something that needs to be fixed by CTDE methods \cite{amato2024introduction}. In this case it is an advantage. By creating a moving target, IPPO prevents the agent from settling into an ``arbitrary handshake" preventing overfitting to its training partners. RPT does this same regularization explicitly through swapping out teammate policies directly and while it is able to reach comparable performance to IPPO, it takes three times as long because policies are manually rotated.

Shared PPO performed poorly in training and in evaluation with teammate policies. The HeMAC problem is complex and stochastic. Because of that, shared-parameter PPO seemed to struggle with handling the multiple different types of agents. The low performance of Shared PPO in ZSC indicates that it has difficulty with the arbitrary handshake problem due to having mirrored teammates. With a mirrored teammate it becomes easy for the agent to develop an idiosyncratic behavior that allows it to only work with certain teammates, creating a difficult challenge when faced with a ZSC environment.

For our evaluations the HeMAC environment served as an excellent testbed. The environment featured distinct roles which necessitated complex coordination between agents. However as mentioned previously the environment had large differences in reward per episode due to the stochastic nature of the target and agent starting location. This created a challenge for the agents, as the reward differences made credit assignment much more difficult.

Because of this, IPPO's performance compared to RPT suggests that decentralized learning may be more robust to unseen teammates than previously suggested in the ZSC literature, at least in the evaluated task domain.

\section{Conclusion}
In this work we explore the ability for MARL to adapt to withheld teammate policies. We introduce Rotation Policy Training (RPT) as a training protocol for Heterogeneous ZSC MARL, and show that RPT performs similarly to IPPO in heterogeneous environments, however is not as sample efficient. In addition, we find that IPPO agents are capable at partnering with novel teammates in the evaluated task domain. We find that architectural complexity is not always a necessary prerequisite for successful cooperative MARL, confirming recent findings in other domains \cite{dewitt2020independent}.

MARL is still a space with many open research questions. Future work can explore how pretraining some of the policies in RPT affects performance. It could also explore techniques to reduce and mitigate the variability in reward presented by the HeMAC environment. Moreover, work can explore how increasing the number and diversity of teammate policies in RPT affects its performance in a ZSC environment. Work can also explore evaluating IPPO and RPT against teammates in human-robot teaming domains.

%%%%%%%%%%%%%%%%%%%%%%%%%%%%%%%%%%%%%%%%%%%%%%%%%%%%%%%%%%%%%%%%%%%%%%%%%%%%%%%%

\newpage
\section*{APPENDIX}

\section{Detailed Hyperparameter Settings} % This will be Appendix A

This appendix details the specific hyperparameters used for training the three different policies (PPO, A2C, and DQN) within the Rotating Policy Training (RPT) framework, as well as the independent PPO baseline. Note that the PPO and A2C policies use on-policy memory buffers, while DQN uses a larger replay memory.

\begin{table}[H]
\makebox[\textwidth][!]{ % Forces the contents to center even if wider than the column
        \begin{minipage}{\textwidth}
	\centering
	\caption{Hyperparameters for Trained Reinforcement Learning Policies}
	\label{tab:app_hyperparameters}
	\resizebox{\textwidth}{!}{% Adjusts table size to column width
		\begin{tabular}{|l|c|c|c|}
			\hline
			\textbf{Parameter}              & \textbf{PPO}          & \textbf{A2C}          & \textbf{DQN}          \\
			\hline
			\multicolumn{4}{|l|}{\textbf{General and Network Parameters}}                                           \\
			\hline
			Learning Rate ($\alpha$)        & $1 \times 10^{-4}$    & $1 \times 10^{-4}$    & $5 \times 10^{-5}$    \\
			\hline
			Discount Factor ($\gamma$)      & 0.99                  & 0.99                  & 0.99                  \\
			\hline
			Network Architecture (Policy/Q) & MLP [256, 256]        & MLP [256, 256]        & MLP [256, 256]        \\
			\hline
			State Preprocessor              & RunningStandardScaler & RunningStandardScaler & RunningStandardScaler \\
			\hline
			\multicolumn{4}{|l|}{\textbf{On-Policy (PPO \& A2C) Specific Parameters}}                               \\
			\hline
			Rollout/Memory Size             & 512                   & 512                   & -                     \\
			\hline
			GAE Lambda ($\lambda$)          & 0.95                  & 0.95                  & -                     \\
			\hline
			Gradient Norm Clip              & 0.4                   & 0.4                   & -                     \\
			\hline
			Entropy Coefficient (Initial)   & 0.015                 & 0.015                 & -                     \\
			\hline
			Number of Learning Epochs       & 3                     & -                     & -                     \\
			\hline
			Mini Batch Size                 & 64                    & 64                    & -                     \\
			\hline
			PPO Clip Parameter ($\epsilon$) & 0.2                   & -                     & -                     \\
			\hline
			\multicolumn{4}{|l|}{\textbf{Off-Policy (DQN) Specific Parameters}}                                     \\
			\hline
			Replay Buffer Size              & -                     & -                     & 10,000                \\
			\hline
			Initial $\epsilon$              & -                     & -                     & 0.7                   \\
			\hline
			$\epsilon$ Timesteps            & -                     & -                     & 600,000               \\
			\hline
			Learning Starts At              & -                     & -                     & 16,000 steps          \\
			\hline
			Random Timesteps                & -                     & -                     & 1,024                 \\
			\hline
		\end{tabular}
	} % End of \resizebox
    \end{minipage}
    }
\end{table} % This will be Table A.1

\end{document}